%% file: main.tex
\documentclass{article}

\usepackage{ijcai24}

\usepackage{times}
\usepackage{soul}
\usepackage{url}
\usepackage[hidelinks]{hyperref}
\usepackage[utf8]{inputenc}
\usepackage[small]{caption}
\usepackage{graphicx}
\usepackage{amsmath}
\usepackage{amsthm}
\usepackage{booktabs}
\usepackage[switch]{lineno}
\usepackage{amsfonts}
\usepackage{multirow}
\usepackage{graphicx}
\usepackage{hyperref}
\usepackage{amssymb}
\usepackage{algpseudocode}
\usepackage{setspace}
\usepackage[linesnumbered,ruled,vlined]{algorithm2e}
\usepackage{xcolor}

\usepackage{amssymb}
\usepackage{bbding}
\usepackage{pifont}
\usepackage{wasysym}
\usepackage{utfsym}
\usepackage{fontawesome}
\usepackage{xcolor} 
\definecolor{CustomGreen}{RGB}{251,24,11} 
\definecolor{CustomRed}{RGB}{7,153,6}
\usepackage{makecell} 

\input{defs}

\title{Balance-aware Sequence Sampling Makes Multi-modal Learning Better}

\author{
    Zhi-Hao Guan
    \affiliations
    Nanjing University of Science and Technology
    \emails
    zhguan@njust.edu.cn
}

\begin{document}

\maketitle

\begin{abstract}
To address the modality imbalance caused by data heterogeneity, existing multi-modal learning (MML) approaches primarily focus on balancing this difference from the perspective of optimization objectives. However, almost all existing methods ignore the impact of sample sequences, i.e., an inappropriate training order tends to trigger learning bias in the model, further exacerbating modality imbalance. In this paper, we propose \underline{B}alance-aware \underline{S}equence \underline{S}ampling (BSS) to enhance the robustness of MML. Specifically,  we first define a multi-perspective measurer to evaluate the balance degree of each sample. 
Via the evaluation, we employ a heuristic scheduler based on curriculum learning (CL) that incrementally provides training subsets, progressing from balanced to imbalanced samples to rebalance MML. Moreover, considering that sample balance may evolve as the model capability increases, we propose a learning-based probabilistic sampling method to dynamically update the training sequences at the epoch level, further improving MML performance.
Extensive experiments on widely used datasets demonstrate the superiority of our method compared with state-of-the-art (SOTA) MML approaches.
\end{abstract}


\section{Introduction}
Multi-modal learning is emerging as a popular research area in artificial intelligence across various scenarios~\cite{r1,r5}, including autonomous vehicles~\cite{r2}, sentiment recognition~\cite{r3}, and information retrieval~\cite{r4}. They have become the main impetus force for improving performance in these tasks through jointly integrating information from diverse sensors.

\begin{figure}[t]
        \includegraphics[width=\linewidth]{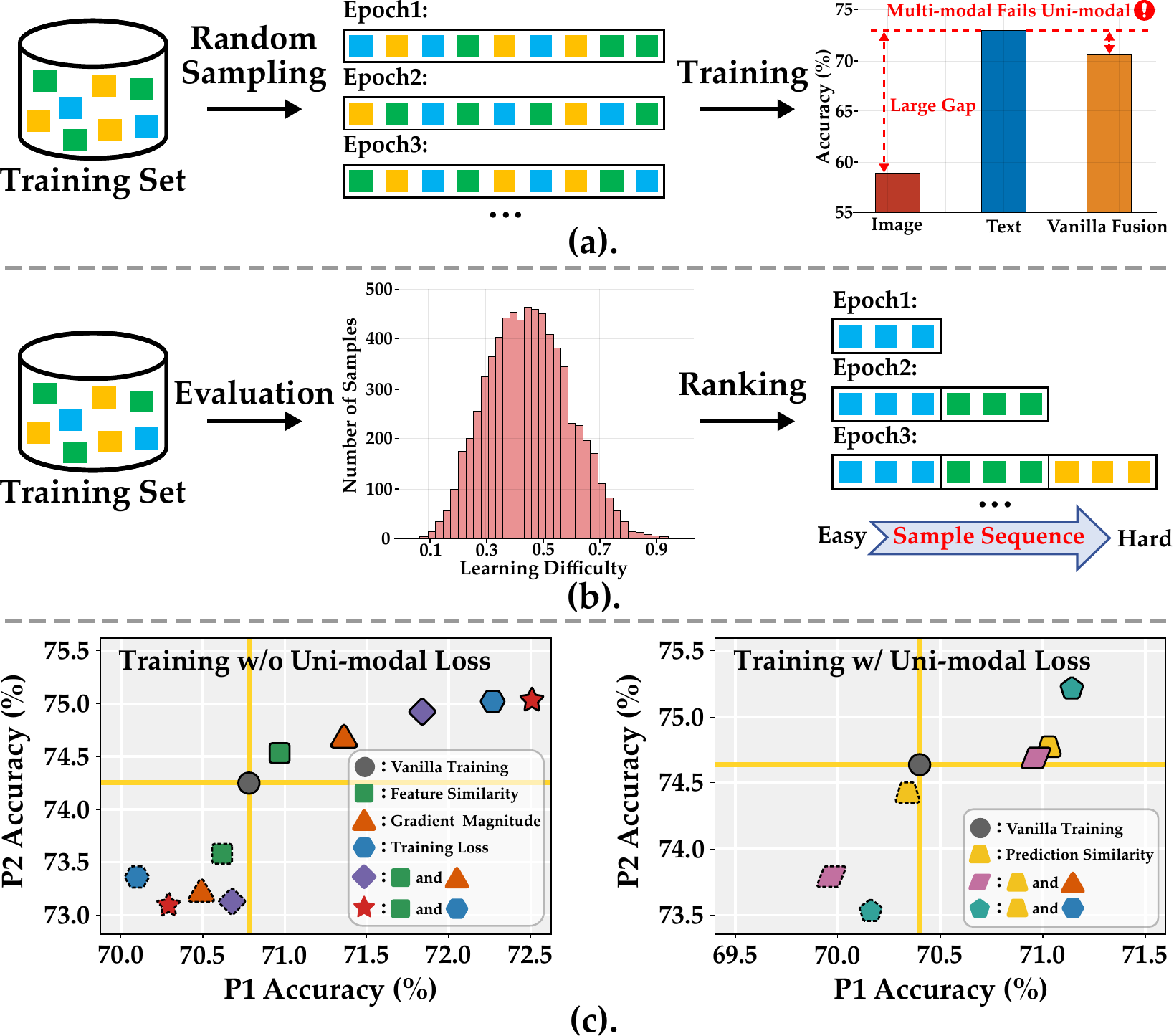}
        \captionsetup{font=normalsize} 
        \captionsetup{width=1\linewidth} 
	\caption{A motivating example of sequence sampling. (a). Traditional vanilla training. It shows that the multi-modal performance fails to outperform the best uni-modal counterpart. (b). Curriculum learning (CL) via training sequence sampling. (c). Comparison of different training paradigms. The results show that CL outperforms the baseline (vanilla training), while anti-CL is inferior to it.}\label{fig:f1}
\end{figure}

Although modalities depict the same concept from different perspectives, each has its unique form~\cite{r6}. For instance, text data is typically represented as word embeddings, while image data consists of pixels. This inherent heterogeneity endows each modality with distinct properties, such as convergence speed~\cite{r7}. As a result, the weak modality (i.e., the slower-converging modality) tends to be underfitted during joint training, leading to modality imbalance, and may even cause the multi-modal model to fail its best uni-modal counterpart~\cite{r8}.

Recently, many impressive studies have been proposed to address the modality imbalance problem from various perspectives~\cite{r10,r11,r12,r13}. Considering the inherent modal differences, a straightforward idea is to manually control the optimization process between strong and weak modalities to realize rebalancing, such as learning rate adjustment~\cite{r14} and gradient modulation~\cite{r15,r7}. Other approaches attempt to facilitate multi-modal learning from the perspective of neural architecture. For instance, UMT~\cite{r16} leverages pre-trained knowledge gained from teacher models to boost uni-modal performance, while AVSlowFast~\cite{r17} randomly removes the network pathway of strong modality with a certain probability.

Although the aforementioned methods can facilitate MML performance to some extent, these solutions address modality imbalance from the perspective of learning objectives while ignoring a key factor, the impact of sample sequences. Since the standard training paradigm is in the form of random data shuffling, each sample has an equal probability of being selected. As shown in Figure~\ref{fig:f1} (a), this inevitably leads to the introduction of imbalanced samples during the early training stages, which can further exacerbate modality imbalance and ultimately degrade overall performance. To support this viewpoint, we conduct a toy experiment on Twitter2015 dataset to investigate the relationship between different sample sequences and MML performance. Inspired by curriculum learning (CL)~\cite{r18,r22}, we first evaluate the learning difficulty of sample pairs based on both information and correlation criteria (e.g., training loss and feature-level similarity), and rank them to construct new sample sequences accordingly, as illustrated in Figure~\ref{fig:f1} (b). We present the comparison results with random sampling in Figure~\ref{fig:f1} (c), where solid outlines indicate CL and dashed outlines indicate anti-CL (i.e., learning from imbalanced to balanced samples). Please note that P1 and P2 refer to the phases when the training samples first reach the entire dataset and the final phase, respectively. To ensure a fair comparison, the size of the training samples in each phase is kept consistent across different approaches. From Figure~\ref{fig:f1} (c), we observe an interesting phenomenon: CL effectively boosts MML performance, while anti-CL tends to have a suppressive effect. This means that MML can benefit from a well-structured sample sequence. The experiment inspires us that by introducing balanced samples during the early training stages, a multi-modal model can alleviate modality imbalance through robust representations, thereby enhancing the overall MML performance.


Based on our findings, in this paper, we attempt to mitigate the modality imbalance problem by adjusting the sample sequences, a training paradigm that provides suitable training subsets to the model at different stages. Concretely, we first design a multi-perspective difficulty measurer to evaluate the balance degree of each data point, which is based on training loss and prediction consistency from both information and correlation criteria. Via sample evaluation, we propose a heuristic and a learning-based sequence sampling method. The former method typically utilizes a fixed learning strategy to construct sample sequences for different training stages. Meanwhile, we propose a more flexible method to dynamically reconstruct training sequences by assigning sampling probabilities to each data point, further enhancing MML performance. To sum up, our contributions are outlined as follows:
\begin{itemize}
    \item To the best of our knowledge, this work is the first attempt to mitigate the modality imbalance problem from the perspective of multi-modal sample sequence.
    \item We propose a heuristic and a learning-based sampling method based on the defined balance score to dynamically adjust the training sequences. Meanwhile, our method can serve as a plug-and-play tool for various MML approaches.
    \item Extensive experiments demonstrate that our proposed method can outperform other baselines and achieve state-of-the-art performance across widely used datasets.
\end{itemize}

\section{Related Work}


\subsection{Imbalanced Multi-modal Learning}
Recent research~\cite{r7,r27} has observed that many multi-modal models fail to outperform the best uni-modal counterpart. This phenomenon is caused by modality imbalance~\cite{r9,r19}, which indicates that each modality cannot be fully learned since there exists inhibition between them. Considering the existence of strong modality and weak modality, some representative~\cite{r8,r15} methods focus on balancing the optimization of individual modalities. In particular, OGM~\cite{r7} proposes on-the-fly gradient modulation technique, which adaptively adjusts the optimization process for each modality by monitoring the discrepancy in their contributions to the learning objective. PMR~\cite{r15} utilizes the prototypes to control updating direction for better uni-modal performance. Other studies~\cite{r16,r20} attempt to boost MML performance by introducing supplementary modules. For instance, UMT~\cite{r16} trains the multi-modal model with knowledge distillation~\cite{r21} from well-learned teacher encoders to obtain richer uni-modal representations. However, these methods increase model complexity and training costs to some extent. In this paper, from the perspective of input sample sequences, we address modality imbalance by guiding the model to progressively learn training samples in a balanced-to-imbalanced manner without additional modules.

\subsection{Curriculum Learning}
Inspired by the organized learning order of knowledge in human cognitive processes, curriculum learning (CL)~\cite{r18,r23} is proposed as a training paradigm that trains machine learning models from easier samples to hard ones. Since CL guides the model toward a better parameter space, recent studies have harnessed its power in various fields, including large language models~\cite{r24}, action recognition~\cite{r25}, and reinforcement learning~\cite{r26}. 
Typically, a curriculum system consists of two main components: a difficulty measurer that evaluates the learning difficulty of samples, while a scheduler determines when and how to assign harder training subsets.
In this paper, we introduce the idea of CL, but with a very different intention. Via learning from balanced samples first, the multi-modal model provides robust representations for subsequent imbalanced samples, as well as avoiding early-stage optimization dilemmas.

\begin{figure*}[t]
	\centering
	\includegraphics[width=180mm]{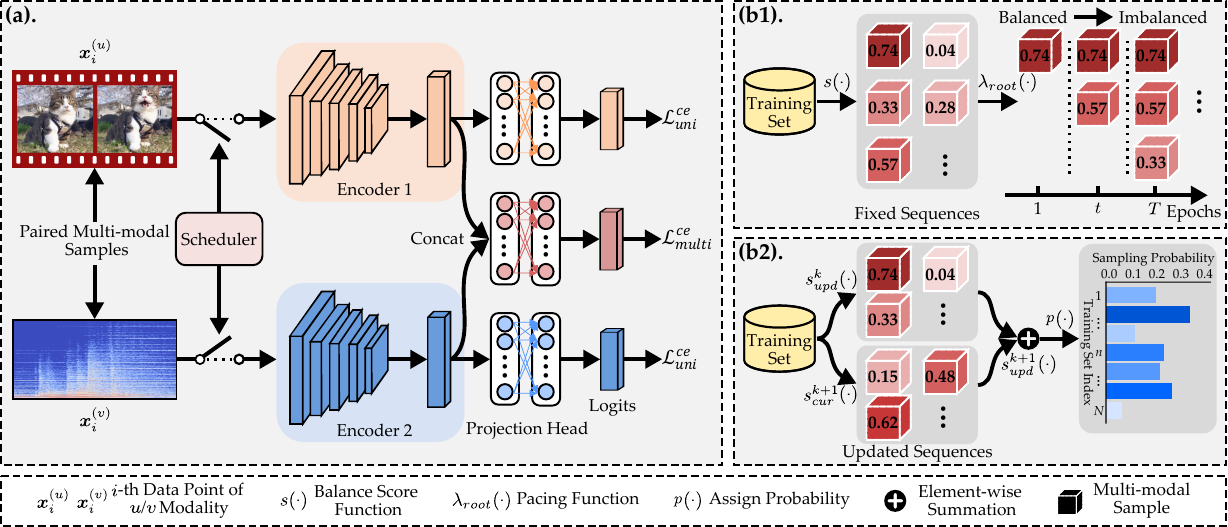}
        \captionsetup{font=normalsize} 
        \captionsetup{width=1\linewidth} 
        \caption{Illustration of BSS method. (a). Multi-modal training framework for learning multi-modal representations. (b1) and (b2). Heuristic and learning-based schedulers for sequence sampling.}\label{fig:f2}
\end{figure*}

\section{Methodology}
In this section, we present our proposed method in detail. The overall architecture is shown in Figure~\ref{fig:f2}, which consists of two main components, i.e., multi-modal training framework and Balance-aware Sequence Sampling (BSS). The former learns multi-modal representations, while the latter rebalances MML using a multi-perspective measurer and two optional schedulers (one heuristic and the other learning-based). 

\subsection{Preliminary}

Without any loss of generality, we consider a multi-modal sample with $u$ and $v$ modalities. Formally, let $\mathcal{D}=\{\X,\Y\}$ denote the dataset, where $\X=\{\x_i^{(u)},\x_i^{(v)}\}_{i=1}^n$ represent $n$ training samples and $\Y=\{\y_i\,|\,\y_i\in\{0,1\}^c\}_{i=1}^n$ is the corresponding category labels with a total of $c$ categories. MML aims to train a model to predict the category label of a given multi-modal sample.

For the MML prediction task, we usually employ a deep neural network to learn the representation of input sample from the original space into feature space. We use $f^{(j)}(\cdot)$ as the feature extractor for $j$-th modality, $j \in \{u,v\}$. Given sample $\x_i^{(j)}$, the feature extraction can be expressed as:
\begin{align}
\e^{(j)}_i=f^{(j)}(\x^{(j)}_i;\theta^{(j)}),
\end{align}
where $\e_i^{(j)}\in\RB^d$ denotes the $d$-dimension feature embedding of $\x^{(j)}_i$, the $\theta^{(j)}$ denotes the learnable parameters of $j$-th encoder. After extracting feature embeddings for all modalities, we adopt a fusion function $g(\cdot)$ to fuse them. Then, we utilize a classifier (i.e., a fully-connected layer) to map the feature embedding into $\RB^c$. This procedure can be formulated as:
\begin{align}
\e_i=g(\e^{(u)}_i,\e^{(v)}_i),\quad\hat{\y}_i=\softmax(\W\e_i+\b).
\end{align}
Here, $\W\in \RB^{c\times D}$, $\b\in \RB^c$ denote the weights and bias of the classifier, respectively, and $D$ denotes the dimension of $\e_i$. Finally, the learning objective is to minimize the multi-modal classification loss, formulated as:
\begin{equation}
\begin{split}
\label{eq:eq3}
\LM_{multi}^{ce}(\X,\Y) = -\frac{1}{n}\sum_{i=1}^n \y^{\top}_i \text{log} \, \hat{\y}_i.
\end{split}
\end{equation}



\subsection{Multi-perspective Measurer}
In the following, we first introduce how to measure the balance degree of a multi-modal sample from the perspectives of correlation and information criteria, and then define the balance score for each sample.\\
\textbf{Correlation Criterion:} Different modalities exhibit inherent cross-modal correlation, as they describe the same concept via diverse representations, capturing complementary information. Although cross-modal correlation can be measured by various aspects, we here only focus on feature similarity and prediction similarity. According to our previous findings in Figure~\ref{fig:f1} (c), the latter outperforms the former. The reason is that prediction similarity is more effective in measuring modality balance, while MML benefits from the supervision of uni-modal predictions~\cite{unismmc}. Formally, given a multi-modal sample $\x_i = \{\x^{(u)}_i,\x^{(v)}_i\}$, the prediction similarity is defined as:
\begin{equation}
\begin{split}
\label{eq:eq4}
\simm(\x^{(u)}_i,\x^{(v)}_i) = \frac{\hat{\y}^{(u)}_i \cdot \hat{\y}^{(v)}_i} {\|\hat{\y}^{(u)}_i\|_2\|\hat{\y}^{(v)}_i\|_2}.
\end{split}
\end{equation}
Here, $\|\cdot\|_2$ denotes $L_2$ norm of the uni-modal predictions. \\
\textbf{Information Criterion:} While prediction similarity reflects the balance between modalities, it does not verify whether the predictions of each modality are correct. In other words, high prediction similarity may still occur even when all modalities produce incorrect predictions. To address this issue, we introduce the label-related training loss as an intuitive metric to further evaluate the learning difficulty of samples, which can be reformulated from Equation~\ref{eq:eq3} as:
\begin{equation}
\begin{split}
\label{eq:eq5}
\LM_{total}(\x^{(u)}_i,\x^{(v)}_i,\y_i) &= (1-\alpha)\LM_{multi}^{ce}(\x^{(u)}_i,\x^{(v)}_i,\y_i) \\
                                        &+ \alpha\sum_{j\in \{u,v\}}\LM_{uni}^{ce}(\x^{(j)}_i,\y_i),
\end{split}
\end{equation}
where $\alpha$ denotes the weighted parameter between two losses. Please note that prediction similarity relies on uni-modal classification results, i.e., the corresponding uni-modal training losses. \\
\textbf{Balance Score:} Overall, the balance score of sample $\x_{i}=\{\x^{(u)}_i,\x^{(v)}_i\}$ can be formulated as the combination of correlation criterion and information criterion:
\begin{equation}
\begin{split}
\label{eq:eq6}
s(\x_i)&=\Norm(\simm(\x^{(u)}_i,\x^{(v)}_i)) \\ 
&- \Norm(\LM_{total}(\x^{(u)}_i,\x^{(v)}_i,\y_i)),
\end{split}
\end{equation}
where $\Norm$($\cdot$) denotes the normalization operation. More definitions of the criteria are presented in the supplementary materials due to space limitations.

\subsection{Training Scheduler}
Merely evaluating the balance score for each sample is insufficient to establish an appropriate training period, as it is also essential to control the presentation of samples from balanced to imbalanced, i.e., the sample sequence for each training epoch. Similar to human education, if teachers impart knowledge from easy to hard within a short span of time, students may become overwhelmed and fail to learn effectively. On the other hand, if teachers present knowledge too slowly, students may lose motivation. In the following, we introduce both a heuristic and a learning-based scheduler to construct sample sequences. \\
\textbf{Heuristic Scheduler:} Inspired by curriculum learning (CL), we rank the training samples from balanced to imbalanced according to the defined balance score, and then employ a pace function~\cite{r28} to control the growth speed of samples. In practice, there exist various pacing functions like baby step~\cite{r18}, liner function~\cite{r22}, and root function~\cite{r29}. However, the effect of existing pacing functions on modality imbalance is not the focus of our work, and comparisons can be found in the supplementary materials. Here, we adopt a widely used root function to achieve this: 
\begin{equation}
\begin{split}
\label{eq:eq7}
\lambda_{root}(t)=\min\left(1,\sqrt{\frac{1-{\lambda}^{2}_0}{T_{grow}} \cdot t + {\lambda}^{2}_0}\right). 
\end{split}
\end{equation}
$\lambda_{root}(t)$ aims to map training epoch number $t$ to a scalar $\lambda\in(0,1]$, which means $\lambda$ proportion of the most balanced samples are available at the $t$-th epoch. This function starts at $\lambda_{root}(0)>0$ and ends at $\lambda_{root}(T_{grow})=1$. $\lambda_0 \in (0,1]$ is the initial proportion of the training samples, and $T_{grow}$ represents the training epoch when this function reaches 1 for the first time.

From  Equation~\ref{eq:eq7}, the pace function can serve as a threshold to continuously expand the sampling space during the training process. The current epoch data $\X_{cur}$ can be randomly sampled from the entire ranked training sequence $\X_{rank}$:
\begin{equation}
\begin{split}
\label{eq:eq8}
\X_{cur}(t)=\RandomSampling(\X_{rank}(t)). 
\end{split}
\end{equation}
Thus, the heuristic scheduler enables the model to focus on balanced samples during the early training stages and gradually broaden the learning scope by incorporating those imbalanced ones. Please note that the sample evaluation is conducted only once before model training, i.e., $\X_{rank}$ is a fixed sequence. \\
\textbf{Learning-based Scheduler:} Despite the simplicity and effectiveness of the heuristic scheduler, there exist several key limitations as follows: 
(1) Without comprehensive trials, it is challenging to find the optimal pacing function for a specific task and its dataset;
(2) The fixed training sequence is inflexible, which may to some extent neglect the feedback from the current model, potentially leading to suboptimal MML performance;
(3) The hyper-parameters of the pacing function, such as $\lambda_0$ and $T_{grow}$ in Equation~\ref{eq:eq7}, are difficult to determine.
Therefore, we propose a learning-based scheduler that flexibly addresses the above limitations. This scheduler reconstructs the dynamic sequence by learning a sampling probability for each data, which takes into account both the balance of past and current samples. 

Specifically, we refer to the current balance score $s_{cur}(\x_i)$ in Equation~\ref{eq:eq6} and update it in a certain epoch interval, use $E$ for short. Then, the $k$+1-th balance score can be represented as:
\begin{equation}
\begin{split}
\label{eq:eq9}
s_{upd}^{k+1}(\x_i) =
\begin{cases}
    s_{cur}^{k+1}(\x_i), & \text{if} \; k=0, \\
    (1-\beta)s_{upd}^{k}(\x_i)+\beta s_{cur}^{k+1}(\x_i), & \text{otherwise},
\end{cases}
\end{split}
\end{equation}
where $k=\lfloor t/E \rfloor$, $t$ denotes the $t$-th epoch, $\beta$ is an adjustment parameter, and $s_{cur}^{1}$ is the balance score obtained before model training, i.e., the first evaluation results.

According to the updated balance score, the learned sampling probability $p$ for each data $\x_i$ in $t$-th epoch can be calculated by softmax operation:
\begin{equation}
\begin{split}
\label{eq:eq10}
p(\x_i)= \frac{e^{s_{upd}^{k+1}(\x_i)}}{\sum_{j=1}^n e^{s_{upd}^{k+1}(\x_j)}}.
\end{split}
\end{equation}
Subsequently, the current epoch data $\X_{cur}$ can be sequentially sampled as follows:
\begin{equation}
\begin{split}
\label{eq:eq11}
\X_{cur}(t)=\SequentialSampling(\bigcup_{i=1}^n \{\x_i \sim p(\x_i)\}). 
\end{split}
\end{equation}
Hence, training data with higher sampling probabilities (i.e., balanced ones) are preferentially selected for the mini-batch in each epoch. The overall algorithm of our BSS model is detailed in Algorithm~\ref{algorithm1}, where $scheduler$ represents the chosen training scheduler in practice.\\
\textbf{Disscssion:} Our proposed method aims to alleviate modality imbalance through sequence sampling in a balanced-to-imbalanced learning manner. Thus, our BSS can be integrated as a model-independent pugin into most existing MML approaches.

\begin{algorithm}[t]
\caption{Multi-modal Learning with Balance-aware Sequence Sampling (\textbf{BSS}).}
\label{algorithm1}
\small
\SetKwInOut{Input}{Input}
\SetKwInOut{Output}{Output}
\SetKw{KwBy}{by}
\DontPrintSemicolon
\SetAlgoLined
\SetNoFillComment
\LinesNotNumbered
\Input{Training set $\X_{train}$, category labels $\Y_{train}$.}
\Output{Learned parameters $\theta$ of multi-modal model.}
\textbf{INIT} initialize parameters $\theta^{0}$, maximum epochs $T$, training set for ranking $\X_{rank}=\emptyset$, curriculum period $T_{grow}$, initial proportion $\lambda_0$, epoch interval $E$.\;

\textcolor{blue}{/* Calculate the balance score via measurer. */} \;
\For{\textnormal{each sample} $\x_i$ \textnormal{in} $\X_{train}$}{
    Obtain balance score $s(\x_i)$ based on Equation~\ref{eq:eq6}. \;
    Add $\x_i$ to $\X_{rank}$ in descending order of $s(\x_i)$. \;
}

\textcolor{blue}{/* Train model using sample sequences from scheduler. */} \\
\For{$t = 0$ \KwTo $T-1$}{
    \uIf{$\text{scheduler} == \text{`\textnormal{heuristic}'}$}{
    Calculate the proportion of the training samples $\lambda_{root}(t)$ with Equation~\ref{eq:eq7}. \;
    Obtain current training sequence $\X_{cur}$ from $\X_{rank}$ with Equation~\ref{eq:eq8}. \;
    }
    \ElseIf{$\text{scheduler} == \text{`\textnormal{learning-based}'}$}{
    Update $s(\x_i)$ every $E$ epochs with Equation~\ref{eq:eq9}. \;
    Assign sampling probability $p(\x_i)$ based on $s(\x_i)$ with Equation~\ref{eq:eq10}. \;
    Obtain current training sequence $\X_{cur}$ with Equation~\ref{eq:eq11}.
    }
    Train model with $\X_{cur}$ and update parameters $\theta$. \;
    Update $t=t+1$.
}
\end{algorithm}

\subsection{Model Inference}
After training the modality-specific encoders and individual classifiers, the learned model can be applied to make prediction during inference stage. Following various late fusion strategies~\cite{r12,r9}, we employ a simple weighted combination of logits output from each modality and their fusion:
\begin{equation}
\begin{split}
\label{eq:eq12}
\z_{total}=\z_{multi}+\sum_{j=1}^{m}\z_{uni}^{(j)}.
\end{split}
\end{equation}
Based on Equation~\ref{eq:eq12}, the predicted category $\hat{y}$ for a given unseen multi-modal sample can be represented as:
\begin{align}
\hat{y}=\underset{i}{\mathtt{argmax}} \frac{e^{\z_{total}^i}}{\sum_{j=1}^{c}e^{\z_{total}^j}},
\end{align}
where $c$ represents the number of category labels.

\section{Experiments}

\subsection{Experimental Setup}
\textbf{Datasets:} We validate our proposed method on six widely used datasets, including CREMA-D~\cite{cremad}, Kinetics-Sounds~\cite{ks}, VGGSound~\cite{vggsound}, Twitter2015~\cite{twitter}, Sarcasm~\cite{sarcasm}, and NVGesture~\cite{nvgesture}. Among them, CREMA-D, Kinetics-Sounds, and VGGSound contain both audio and video modalities. CREMA-D includes 7,442 video clips across six emotional categories, with 6,698 clips for training and 744 for testing. Kinetics-Sounds consists of 19,000 clips categorized into 31 distinct actions, split into 15,000 for training, 1,900 for validation, and 1,900 for testing. VGGSound provides 168,618 videos for training and validation, along with 13,954 videos for testing. Moreover, Twitter2015 and Sarcasm datasets involve both image and text modalities. Twitter2015 comprises 5,338 text-image pairs, divided into 3,179 for training, 1,122 for validation, and 1,037 for testing. Sarcasm contains 24,635 text-image combinations, allocated as 19,816 for training, 2,410 for validation, and 2,409 for testing. Lastly, NVGesture features three modalities, i.e., RGB, optical flow (OF), and Depth, with 1,050 samples for training and 482 samples for testing. \\
\textbf{Baselines and Evaluation Metrics:} We conduct a comprehensive comparison of BSS with two types of baselines: vanilla fusion methods and multi-modal rebalance approaches. The former includes seven techniques, such as concatenation (Concat) and summation (Sum). The latter comprises 11 SOTA methods, including OGM~\cite{r7} and MLA~\cite{r12}. For more detailed baseline descriptions, please refer to the supplementary material.

Following~\cite{r7,reconboost}, we utilize accuracy (ACC), mean average precision (MAP), and Macro F1-score (Mac-F1) as evaluation metrics. ACC quantifies the ratio of correct predictions to total predictions. MAP reflects the average precision of all samples, while Mac-F1 calculates the average of F1 scores across all categories.\\
\textbf{Implementation Details:} Following~\cite{r7,r15}, for audio-video datasets, we use ResNet18~\cite{resnet} as the backbone to encode each modality. For text-image datasets, we employ ResNet50 for images and BERT~\cite{bert} for text processing. For the tri-modal dataset NVGesture, we follow the setup of~\cite{r20} and adopt the I3D~\cite{i3d} as uni-modal backbone. To ensure fairness, all methods utilize a consistent backbone during training. The optimizer for the audio-video datasets is stochastic gradient descent (SGD) with a momentum of 0.9 and weight decay of 10$^{-4}$. The initial learning rate is set to 10$^{-2}$ and is reduced by a factor of 10 when the loss saturates. The batch size is set to 64 for CREMA-D and Kinetics-Sounds, 16 for VGGSound, and 2 for NVGesture. For text-image datasets, we employ the Adam optimizer starting with a learning rate of 10$^{-5}$, with a batch size of 64. Furthermore, hyper-parameter $\alpha$ and $\beta$ are set to 0.2 and 0.6. For the training scheduler, the curriculum period $T_{grow}$ and the initial proportion $\lambda_0$ are configured as 40 and 0.1 in the heuristic setting, while the epoch interval $E$ is configured as 5 in the learning-based setting. All models are trained on a single RTX 3090 GPU.

\begin{table*}[t]
    \centering
    \captionsetup{font=normalsize} 
    \captionsetup{width=1\linewidth} 
    \caption{Comparison with SOTA multi-modal learning methods. The best performances are highlighted in bold. The underlining symbol denotes the second-best performance.}
    \label{tab:tab1}
    \setlength{\tabcolsep}{3.5pt} 
    \begin{tabular}{lcccccccccc}
        \toprule
        \multirow{2}{*}{Method} & \multicolumn{2}{c}{CREMA-D} & \multicolumn{2}{c}{Kinetics-Sounds} & \multicolumn{2}{c}{Twitter2015} & \multicolumn{2}{c}{Sarcasm} & \multicolumn{2}{c}{NVGesture}\\
        \cmidrule(lr){2-3} \cmidrule(lr){4-5} \cmidrule(lr){6-7} \cmidrule(lr){8-9} \cmidrule(lr){10-11}
        & ACC\,(\%) & MAP\,(\%) & ACC\,(\%) & MAP\,(\%) & ACC\,(\%) & F1\,(\%) & ACC\,(\%) & F1\,(\%) & ACC\,(\%) & F1\,(\%)\\
        \midrule
        Audio/Text/RGB   & 63.17 & 68.61 & 54.12 & 56.69 & 73.67 & 68.49 & 81.36 & 80.65 & 78.22 & 78.33\\
        Video/Image/OF & 45.83 & 58.79 & 55.62 & 58.37 & 58.63 & 43.33 & 71.81 & 70.73 & 78.63 & 78.65\\
        Depth      & - & - & - & - & - & - & - & - & 81.54 & 81.83\\
        \midrule
        Concat       & 63.31 & 68.31 & 64.55 & 71.31 & 70.11 & 63.86 & 82.86 & 82.43 & 81.33 & 81.47\\
        Affine       & 66.26 & 71.93 & 64.24 & 69.31 & 72.03 & 59.92 & 82.47 & 81.88 & 82.78 & 82.81\\
        Channel      & 66.13 & 71.75 & 63.51 & 68.66 & -     & -     & -     & -     & 81.54 & 81.57  \\
        ML-LSTM      & 62.94 & 64.73 & 63.84 & 69.02 & 70.68 & 65.64 & 82.05 & 70.73 & 83.20 & 83.30\\
        Sum          & 63.44 & 69.08 & 64.97 & 71.03 & 73.12 & 66.61 & 82.94 & 82.47 & 82.99 & 83.05\\
        Weight       & 66.53 & 73.26 & 65.33 & 71.33 & 72.42 & 65.16 & 82.65 & 82.19 & 83.42 & 83.57\\
        ETMC         & 65.86 & 71.34 & 65.67 & 71.19 & 73.96 & 67.39 & 83.69 & 83.23 & 83.61 & 83.69\\
        \midrule
        MSES \textcolor{blue}{\scriptsize{[ACPR'19]}}         & 61.56 & 68.83 & 64.71 & 70.63 & 71.84 & 66.55 & 84.18 & 83.60 & 81.12 & 81.47\\
        OGR-GB \textcolor{blue}{\scriptsize{[CVPR'20]}}      & 64.65 & 84.54 & 67.10 & 71.39 & 74.35 & 68.69 & 83.35 & 82.71 & 82.99 & 83.05\\
        DOMFN \textcolor{blue}{\scriptsize{[MM'22]}}       & 67.34 & 85.72 & 66.25 & 72.44 & 74.45 & 68.57 & 83.56 & 82.62 & - & -\\
        OGM \textcolor{blue}{\scriptsize{[CVPR'22]}}         & 66.94 & 71.73 & 66.06 & 71.44 & 74.92 & 68.74 & 83.23 & 82.66 & - & -\\
        MSLR \textcolor{blue}{\scriptsize{[ACL'22]}}        & 65.46 & 71.38 & 65.91 & 71.96 & 72.52 & 64.39 & 84.23 & 83.69 & 82.86 & 82.92\\
        AGM \textcolor{blue}{\scriptsize{[ICCV'23]}}         & 67.07 & 73.58 & 66.02 & 72.52 & \underline{74.83} & \underline{69.11} & 84.02 & 83.44 & 82.78 & 82.82\\ 
        PMR \textcolor{blue}{\scriptsize{[CVPR'23]}}         & 66.59 & 70.30 & 66.56 & 71.93 & 74.25 & 68.60 & 83.60 & 82.49 & - & -\\
        ReconBoost \textcolor{blue}{\scriptsize{[ICML'24]}}  & 74.84 & 81.24 & 70.85 & 74.24 & 74.42 & 68.34 & 84.37 & 83.17 & 84.13 & \underline{86.32}\\
        MMPareto \textcolor{blue}{\scriptsize{[ICML'24]}}    & 74.87 & 85.35 & 70.00 & 78.50 & 73.58 & 67.29 & 83.48 & 82.48 & 83.82 & 84.24\\
        SMV \textcolor{blue}{\scriptsize{[CVPR'24]}}         & 78.72 & 84.17 & 69.00 & 74.26 & 74.28 & 68.17 & 84.18 & 83.68 & 83.52 & 83.41\\
        MLA \textcolor{blue}{\scriptsize{[CVPR'24]}}         & 79.43 & 85.72 & 70.04 & 74.13 & 73.52 & 67.13 & 84.26 & 83.48 & 83.40 & 83.72\\
        \midrule
        BSS-H       & \underline{80.78} & \underline{87.86} & \underline{72.67} & \underline{78.61} & 74.73             & 68.67             & \underline{84.41} & \underline{83.86} & \underline{85.06} & 85.15\\
        BSS-L      & \textbf{82.80}    & \textbf{88.61}    & \textbf{73.95}    & \textbf{79.43}    & \textbf{75.22}    & \textbf{69.51}    & \textbf{85.01}    & \textbf{84.62} & \textbf{86.72} & \textbf{87.04}\\
        \bottomrule
    \end{tabular}
\end{table*}

\subsection{Comparsion with SOTA MML Baselines}

We conduct comprehensive comparisons to assess the superiority of our proposed method in addressing the imbalanced multi-modal learning problem. The classification performance on all datasets is reported in Table~\ref{tab:tab1} and Table~\ref{tab:tab2}, where ``BSS-H'' and ``BSS-L'' denote the proposed method with heuristic scheduler and learning-based scheduler, respectively. Please note that ``-'' in Table~\ref{tab:tab1} denotes that the corresponding methods are not applicable to datasets with more than two modalities. \\
\textbf{Results on Bi-modal Dataset:} Referring to the first four datasets in Table~\ref{tab:tab1}, we derive the following key observations: (1). Uni-modal performance may outperform multi-modal joint training. For example, the text-modal performance on the Twitter2015 dataset is obviously better than most vanilla fusion methods, indicating an inhibitory relationship between different modalities; (2). Most multi-modal rebalance approaches demonstrate significant improvements over vanilla fusion methods. This phenomenon not only underscores the adverse impact of modality imbalance on performance but also validates the effectiveness of the multi-modal rebalance approach; (3). Compared with all baselines including vanilla fusion methods and multi-modal rebalance approaches, our proposed method achieves the best performance by a large margin across all metrics. It can be observed that BSS-L delivers significant performance improvements on both CREMA-D and Kinetics-Sounds datasets. After sequence sampling, our method surpasses the best baseline (MLA)~\cite{r12} with gains of 3.37\%/2.89\% and 3.91\%/5.30\% in ACC and MAP metrics, respectively. \\
\textbf{Results on Tri-modal Dataset:} In addition, we present a comparison with SOTA baselines on the NVGesture dataset. As shown in the last dataset of Table~\ref{tab:tab1}, unlike multi-modal rebalance approaches limited to scenarios with only two modalities (e.g., OGM and PMR), our method effectively tackles the challenges in scenarios involving more than two modalities and achieves the best performance. \\
\textbf{Results on Large-scale Dataset:} To further evaluate the generality of our method, we conduct experiments on a large-scale dataset, VGGSound. Given the size of the dataset, we only select a few representative baselines for comparison, including OGM, AGM, ReconBoost, MMPareto, SMV, and MLA. The results in Table~\ref{tab:tab2} consistently demonstrate that our BSS-L achieves superior performance.


\begin{table}[h]
    \centering
    \captionsetup{font=normalsize} 
    \captionsetup{width=0.85\linewidth} 
    \caption{Performances on VGGSound dataset.}
    \label{tab:tab2}
    \begin{tabular}{lcc}
        \toprule
        Method  & ACC\,(\%)            & MAP\,(\%)     \\
        \midrule
        OGM        & 48.29             & 49.78 \\
        AGM        & 47.11             & 51.98 \\
        ReconBoost & 50.97             & 53.87 \\
        MMPareto   & 51.25             & 54.73 \\
        SMV        & 50.31             & 53.62 \\
        MLA        & \underline{51.65} & 54.73 \\
        \midrule
        BSS-H     & 51.61          & \underline{55.68} \\
        BSS-L     & \textbf{52.80} & \textbf{56.61} \\
        \bottomrule
    \end{tabular}
\end{table} 

\subsection{Ablation Study}
We conduct more ablation studies to verify the effectiveness of using different criteria for sample evaluation, i.e., uni-modal prediction similarity (PreSim) and training loss (Loss). Table~\ref{tab:tab3} records the results under the learning-based setting, which reveal that: 
(1). Vanilla joint training may exacerbate modality imbalance. For instance, when the video modality converges, the audio modality remains insufficiently trained, leading to a significant gap between the two modalities (4.66\%/6.41\% in ACC/MAP) compared with other variants;
(2). Both uni-modal prediction similarity and training loss, when employed as balance scores for sequence sampling, can boost classification performance. This is predictable, as prioritizing balanced samples helps reduce the gap between modalities, facilitating both uni-modal and multi-modal learning processes;
(3). By integrating both criteria, BSS-L exhibits the best performance, demonstrating its effectiveness in modality imbalance scenarios.

\begin{table}[h]
    \centering
    \captionsetup{font=normalsize} 
    \captionsetup{width=1\linewidth} 
    \caption{Ablation study on Kinetics-Sounds dataset under the learning-based setting.}
    \label{tab:tab3}
    \setlength{\tabcolsep}{5pt} 
    \begin{tabular}{ccccc}
        \toprule
        \multicolumn{2}{c}{Criterion} & \multicolumn{3}{c}{ACC\,(\%) \;/\; MAP\,(\%)} \\
        \cmidrule(lr){1-2} \cmidrule(lr){3-5}
        PreSim       & Loss          & Audio               & Video            & Multi  \\
        \midrule
        \textcolor{CustomRed}{\ding{55}} & \textcolor{CustomRed}{\ding{55}}          & 49.37/51.07                      & 54.03/57.48                           & 70.44/76.62   \\
        \textcolor{CustomRed}{\ding{55}} & \textcolor{CustomGreen}{\ding{51}}        & 52.11/\underline{54.40}          & 54.23/57.91                           & 72.44/\underline{79.41}   \\
        \textcolor{CustomGreen}{\ding{51}} & \textcolor{CustomRed}{\ding{55}}        & \underline{52.38}/54.32          & \textbf{54.93}/\textbf{58.52}      & \underline{73.25}/78.98   \\
        \textcolor{CustomGreen}{\ding{51}} & \textcolor{CustomGreen}{\ding{51}}      & \textbf{52.73}/\textbf{54.43}    & \underline{54.74}/\underline{58.46}      & \textbf{73.95}/\textbf{79.43}  \\
        \bottomrule
    \end{tabular}
\end{table}

\subsection{Further Analysis}
\begin{figure}[h]
        \includegraphics[width=\linewidth]{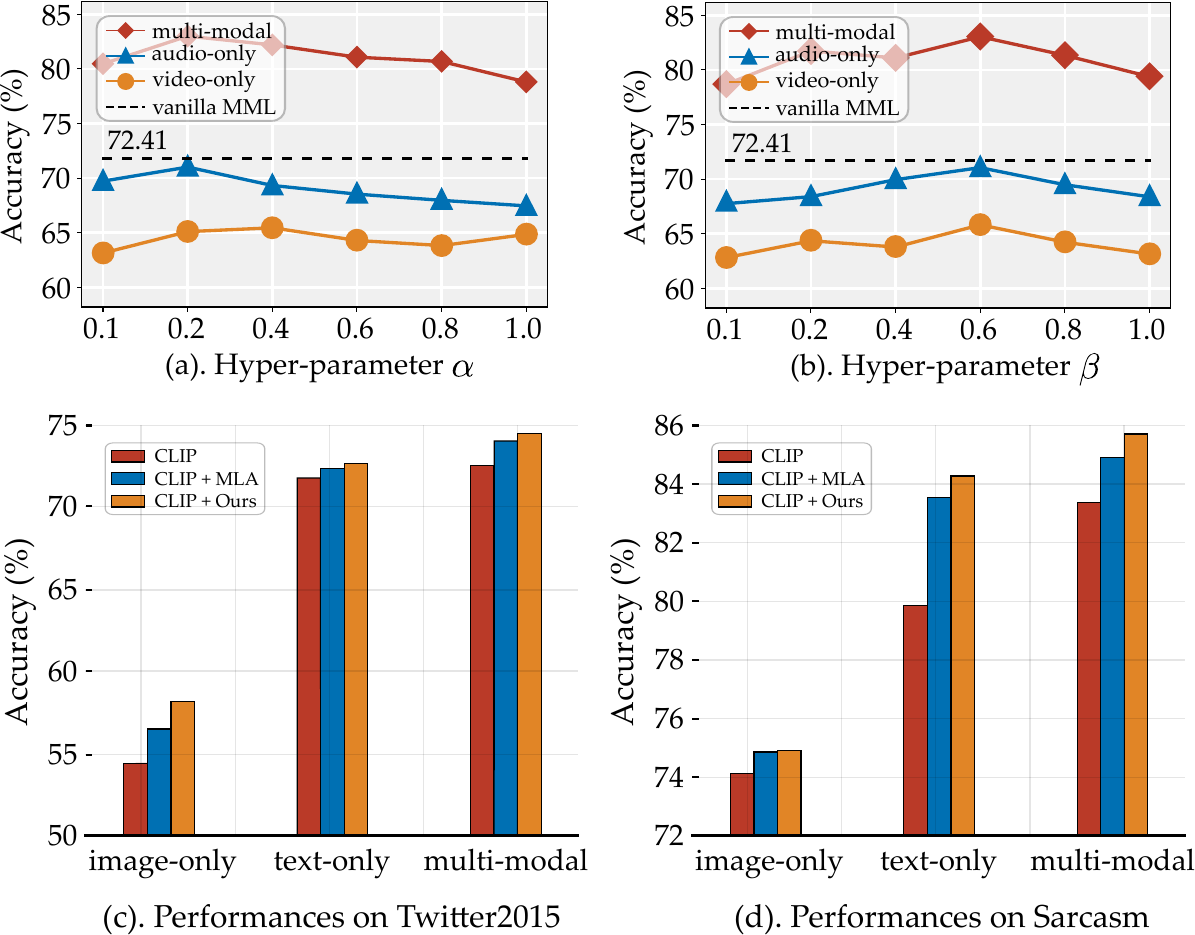}
        \captionsetup{font=normalsize} 
        \captionsetup{width=1\linewidth} 
	\caption{(a) and (b). Comparison with hyper-parameters on CREMA-D dataset. (c) and (d). Robust performance achieved by using the CLIP pre-trained model as encoders.}\label{fig:f3}
\end{figure}

\begin{figure}[h]
	\centering
	\includegraphics[width=\linewidth]{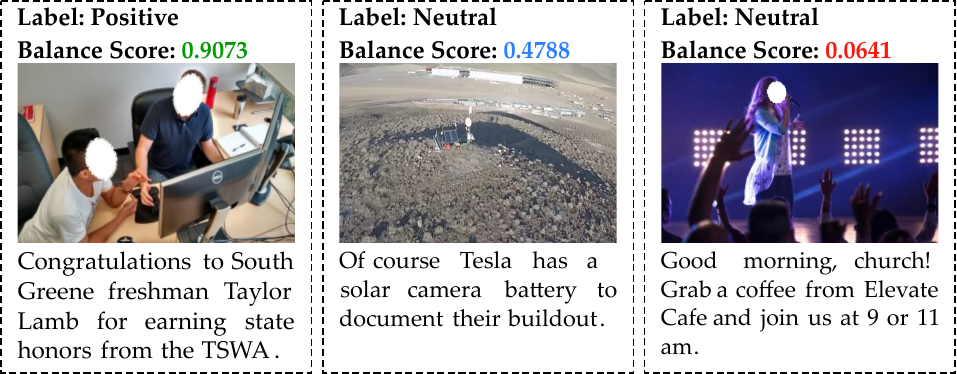}
        \captionsetup{font=normalsize} 
        \captionsetup{width=1\linewidth} 
        \caption{Qualitative results of sample evaluation. We show some representative samples selected from different segments (early, middle, and late) of the evaluation sequence.
        }
        \label{fig:f4}
\end{figure}

\textbf{Sensitivity to Hyper-Parameters:} In calibrating our proposed method, we identify two hyper-parameters: $\alpha$ in Equation~\ref{eq:eq5} and $\beta$ in Equation~\ref{eq:eq9}, determining the strength for balancing classification loss and regulating the balance score, respectively. Figure~\ref{fig:f3} (a) depicts the performance of different $\alpha$. With the increase of $\alpha$, the accuracy of our method first increases and then decreases. This shows that proper uni-modal learning has a promoting effect, but over-considering uni-modal optimization may hinder multi-modal interactions. From Figure~\ref{fig:f3} (b), we can find that the performance is marginally affected by $\beta$, suggesting the insensitivity of our method to hyper-parameters. Despite some fluctuations, our method still demonstrates excellent effectiveness, i.e., being consistently better than vanilla multi-modal learning.\\
\textbf{Robustness of the Pre-trained Model:} We further explore the robustness of the large multi-modal pre-trained model on text-image datasets. We replace each modality encoder with the corresponding encoder pre-trained by CLIP~\cite{clip} and fine-tune the model. The results are shown in Figure~\ref{fig:f3} (c) and (d), where ``CLIP+MLA'' and ``CLIP+Ours'' present that we apply the MLA's and our approach, respectively. From Figure~\ref{fig:f3} (c) and (d), we can draw the following observations: (1). Both CLIP+MLA and CLIP+Ours can outperform CLIP in all cases; (2). Via sequence sampling, the performance of our method is better than that of MLA.\\
\textbf{Case Study:} We examine whether our method can effectively distinguish the balanced and imbalanced samples from a randomly ordered sequence. From the representative samples in Figure~\ref{fig:f4}, we observe that balanced samples exhibit strong semantic consistency between modalities, reflected by high balance scores, whereas imbalanced samples typically display weak semantic connections or irrelevant information.


\section{Conclusion}
In this paper, we propose a novel multi-modal learning method, called balance-aware sequence sampling (BSS). By defining a multi-perspective measurer, we evaluate the balance of each training sample. Via the evaluation, we design a heuristic and a learning-based scheduler to construct sample sequences for the model at different training stages. Thus, BSS alleviates modality imbalance through sequence sampling in a balanced-to-imbalanced learning strategy, further boosting MML performance. Moreover, BSS can be integrated as a model-independent plugin into most existing MML approaches. Extensive experiments on widely used datasets validate the superiority of BSS over SOTA baselines.

\bibliographystyle{named}
\bibliography{ijcai24}

\end{document}


\maketitle

\appendix

\section{Criterion Definition}
We summarize the criteria defined for evaluating training samples in Table~\ref{tab:tab1}.

\begin{table*}[t]
    \centering
    \captionsetup{font=normalsize} 
    \captionsetup{width=1\linewidth} 
    \caption{Notation and criteria definition.}
    \label{tab:tab1}
    \setlength{\tabcolsep}{20pt} 
    \begin{tabular}{l|l}
        \toprule
        Notation/Criteria & Description \\
        \midrule
        $\x_i=\{\x_i^{(u)},\x_i^{(v)}\}$      & $i$-th sample pair with $u$ and $v$ modalities. \\
        \midrule
        $\y_i/\hat{\y_i}$                      & Category label/Predicition of $i$-th sample. \\
        \midrule
        $\e_i^{(j)}$                          & Feature embedding of $\x_i^{(j)}$. \\   
        \midrule
        $\Norm(\cdot)$                        & Normalization operation.\\    
        \midrule
        $\simm(\e^{(u)}_i,\e^{(v)}_i) = \frac{\e^{(u)}_i \cdot \e^{(v)}_i} {\|\e^{(u)}_i\|_2\|\e^{(v)}_i\|_2}$                               & Feature similarity.\\
        \midrule
        $\simm(\hat{\y}^{(u)}_i,\hat{\y}^{(v)}_i) = \frac{\hat{\y}^{(u)}_i \cdot \hat{\y}^{(v)}_i} {\|\hat{\y}^{(u)}_i\|_2\|\hat{\y}^{(v)}_i\|_2}$                               & Prediction similarity.\\
        \midrule
        $\LM^{ce}(\x_i,\y_i) = -\y^{\top}_i \text{log} \, \hat{\y}_i.$                                   & Training loss.\\
        \midrule
        $\|\nabla_\theta \LM^{ce}\|_2 = \sqrt{\sum_{i} \left( \frac{\partial \mathcal{L}^{ce}}{\partial \theta_i} \right)^2}$                               & Gradient magnitude.\\
        \midrule
        $\Norm(\simm(\e^{(u)}_i,\e^{(v)}_i)) - \Norm(\|\nabla_\theta \LM^{ce}\|_2)$                                & Feature similarity \& Gradient magnitude.\\
        \midrule
        $\Norm(\simm(\e^{(u)}_i,\e^{(v)}_i)) - \Norm(\LM^{ce}(\x_i,\y_i))$                                & Feature similarity \& Training loss.\\
        \midrule
        $\Norm(\simm(\hat{\y}^{(u)}_i,\hat{\y}^{(v)}_i)) - \Norm(\|\nabla_\theta \LM^{ce}\|_2)$           & Prediction similarity \& Gradient magnitude.\\
        \midrule
        $\Norm(\simm(\hat{\y}^{(u)}_i,\hat{\y}^{(v)}_i)) - \Norm(\LM^{ce}(\x_i,\y_i))$                    & Prediction similarity \& Training loss.\\
        \bottomrule
    \end{tabular}
\end{table*}

\section{Dataset Overview and Baseline Details}

\subsection{Dataset Overview}
\textbf{CREMA-D.} CREMA-D~\cite{cremad} is an audio-visual dataset designed for emotion recognition research. It comprises 7,442 short video clips, each 2$\sim$3 seconds long, featuring 91 actors delivering brief phrases. The dataset is divided into 6,698 clips for training and 744 clips for testing, encompassing six distinct emotional categories: angry, happy, sad, neutral, disgust, and fear.
\\
\textbf{Kinetics-Sounds.} Kinetics-Sounds~\cite{ks} is a dataset bridging computer vision and audio processing. It contains 19,000 video clips, each approximately 10$\sim$seconds long, sourced from YouTube. The dataset is organized into 15,000 clips for training, 1,900 for validation, and 1,900 for testing, covering 31 action categories.
\\
\textbf{VGGSound.} VGGSound~\cite{vggsound} is a widely-used dataset for audio-visual recognition, comprising nearly 200K 10$\sim$second video clips across more than 300 categories. In our experiments, the dataset includes 168,618 videos for training and validation, and 13,954 videos for testing.
\\
\textbf{Twitter2015.} Twitter2015~\cite{twitter} is a dataset created for sentiment analysis and is divided into a training set of 3,197 pairs, a validation set of 1,122 pairs, and a testing set of 1,037 pairs. It comprises tweets collected from Twitter, labeled with three sentiments: positive, negative, and neutral.
\\
\textbf{Sarcasm.} Sarcasm~\cite{sarcasm} is a unique dataset designed for sarcasm detection tasks, providing a large amount of image-text pairs with clear sarcasm annotations. The data source comes from social media, news comments, and conversations. It contains 19,816 training pairs, 2,410 validation pairs, and 2,409 testing pairs with two labels: sarcastic and non-sarcastic.
\\
\textbf{NVGesture.} NVGesture~\cite{nvgesture} is a tri-modal dataset for gesture recognition, featuring 25 categories and encompassing RGB, optical flow (OF), and Depth modalities. It contains a total of 1,532 dynamic hand gestures, with 1,050 samples for training and 482 for testing.

\subsection{Baseline Details}
\textbf{MSES.} MSES~\cite{mses} introduces a multi-modal machine learning approach using multi-task learning to mitigate overfitting caused by a shared learning rate across different modalities. The method simultaneously trains individual classifiers for each modality and a combined multi-modal classifier, with early stopping applied separately to each modality to avoid overfitting.
\\
\textbf{OGR-GB.} OGR-GB~\cite{r8} explores why multi-modal networks often underperform compared to uni-modal ones, attributing it to overfitting caused by increased capacity and varying generalization rates across modalities. To solve these issues, OGR-GB introduces Gradient-Blending, a method that optimally combines modalities according to their overfitting tendencies.
\\
\textbf{DOMFN.} DOMFN~\cite{domfn} adaptively fuses uni-modal and multi-modal predictions based on a divergence penalty score. DOMFN selectively performs fusion when cross-modal divergence is low, improving performance and providing better results than uni-modal approaches.
\\
\textbf{OGM.} OGM~\cite{r7} dynamically adjusts the optimization of each modality according to its impact on the learning objective. It also incorporates dynamic Gaussian noise to mitigate generalization problems, resulting in notable improvements over baseline methods.
\\
\textbf{MSLR.} MSLR~\cite{r14} creates late-fusion multi-modal models from fine-tuned uni-modal ones. By exploring three strategies for assigning learning rates to each modality, MSLR outperforms global learning rates in various tasks, enabling more effective learning of individual modalities.
\\
\textbf{AGM.} AGM~\cite{agm} enhances the performance of multi-modal models with adaptive gradient modulation. Additionally, AGM proposes a new metric to quantitatively measure the competition strength between modalities, further improving MML performance.
\\
\textbf{PMR.} PMR~\cite{r15} employs prototypes to build non-parametric classifiers for uni-modal evaluation and accelerates the slow-learning modality by enhancing clustering towards these prototypes. It also introduces prototype-based entropy regularization in early training to prevent premature convergence caused by the dominant modality.
\\
\textbf{ReconBoost.} ReconBoost~\cite{reconboost} employs a modality-alternating learning paradigm to handle competition with historical models. Unlike classic Gradient-Boosting, it retains only the most recent model per modality, avoiding overfitting from ensembling strong learners.
\\
\textbf{MMPareto.} MMPareto~\cite{mmpareto} analyzes Pareto integration in the multimodal context and ensures a unified gradient direction with enhanced magnitude to improve generalization, providing effective unimodal assistance.
\\
\textbf{SMV.} SMV~\cite{smv} introduces a sample-level modality valuation metric to assess each modality's contribution. It reveals sample-level discrepancies, beyond global differences, and improves modality cooperation by boosting the discriminative ability of low-contributing modalities.
\\
\textbf{MLA.} MLA~\cite{r12} redefines joint multimodal learning by alternating unimodal learning to reduce modality interferences, while capturing cross-modal interactions through a shared head optimized across modalities. During inference, MLA integrates multimodal data using a test-time uncertainty-based fusion mechanism.

\begin{figure*}[t]
	\centering
	\includegraphics[width=180mm]{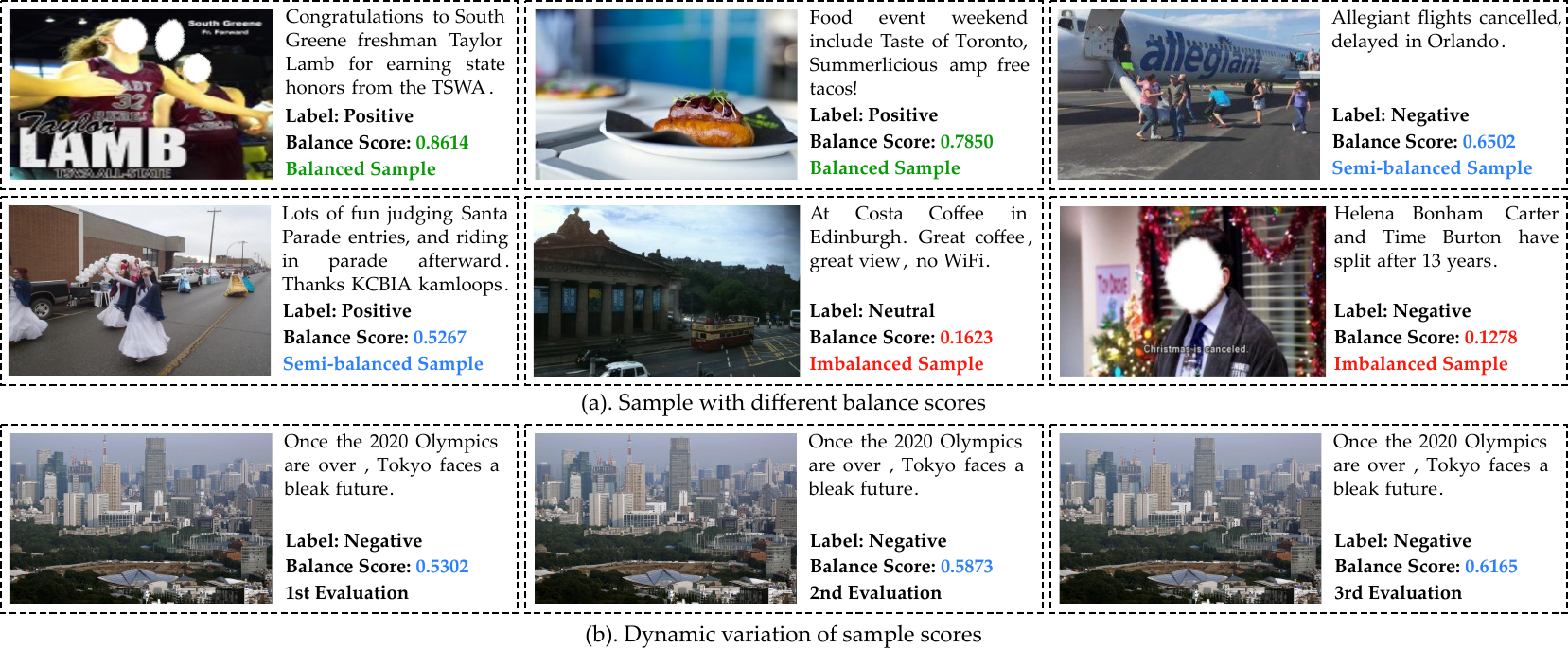}
        \captionsetup{font=normalsize} 
        \captionsetup{width=1\linewidth} 
        \caption{Qualitative results of sample evaluation. (a). Some representative samples selected from different segments on the Twitter2015 dataset, designated as balanced, semi-balanced, and imbalanced. (b). Dynamic variation of sample scores with the learning-based scheduler.}\label{fig:f1}
\end{figure*}

\section{Additional Experiments}

\subsection{Case Study}
In this subsection, we present more sample evaluation cases.
Figure~\ref{fig:f1} (a) presents some representative samples selected from different segments of the training sequence on Twitter2015 dataset. We observe the following: 
(1). Samples from the early segment of the training sequence, designated as balanced sample, exhibit strong semantic consistency between the image and text. For instance, balance scores such as 0.8614 and 0.7850 reflect a clear and coherent relationship between modalities, making these pairs well-suited for initial training stages; 
(2). In contrast, samples from the middle segment, designated as semi-balanced sample, are ambiguous. These pairs reflect partial similarity, where the image and text share some relevance but lack full coherence; 
(3). On the other hand, imbalanced samples typically exhibit weak or nonexistent semantic connections. For instance, the sample with a balance score of only 0.1278 highlights a clear semantic mismatch, where the image and text convey conflicting or irrelevant information. 
Furthermore, we present the variation of balance scores for the same sample under the learning-based scheduler. As shown in Figure~\ref{fig:f1} (b), with the improvement of model ability, our proposed method gradually adapts to more challenging samples, effectively balancing the stronger and weaker modalities.

\subsection{Toy Experiment on CREMA-D Dataset}
To further validate our motivation, i.e., training sequences affect imbalanced multi-modal learning, we conduct more toy experiments on the CREMA-D dataset. For simplicity, we use the root function as the scheduler, setting the curriculum period $T_{grow}$ and the initial sample proportion $\lambda_0$ to 40 and 0.1, respectively. From Table~\ref{tab:tab2}, we observe conclusions consistent with those from the Twitter2015 dataset: (1). Curriculum learning (CL) outperforms vanilla training, whereas anti-CL performs worse; (2). Using prediction similarity and training loss as evaluation criteria leads to the best performance under CL setting. Note that ``*'' indicates the use of cross-entropy classification loss during joint training.

\begin{table*}[t]
    \centering
    \captionsetup{font=normalsize} 
    \captionsetup{width=1\linewidth} 
    \caption{Comparison of training paradigms across different evaluation criteria.}
    \label{tab:tab2}
    \setlength{\tabcolsep}{3.5pt} 
    \begin{tabular}{lcccc}
        \toprule
        \multirow{2}{*}{Evaluation Criteria} & \multicolumn{4}{c}{ACC\,(\%)}\\
        \cmidrule(lr){2-5}
                                                         & P1 (CL) & P2 (CL) & P1 (Anti-CL) & P2 (Anti-CL)\\
        \midrule
        Vanilla Training                                 & 66.83 (70.59*) & 70.89 (74.88*) & 66.83 (70.59*)  & 70.89 (74.88*)\\
        Feature Similarity                               & 69.92\textcolor{CustomRed}{\scriptsize{(+3.09)}}  & 73.37\textcolor{CustomRed}{\scriptsize{(+2.48)}}  & 63.60\textcolor{CustomGreen}{\scriptsize{(-3.23)}}  & 70.14\textcolor{CustomGreen}{\scriptsize{(-0.75)}}\\
        Prediction Similarity*                           & 73.54\textcolor{CustomRed}{\scriptsize{(+2.95)}}  & 77.12\textcolor{CustomRed}{\scriptsize{(+2.24)}}  & 65.74\textcolor{CustomGreen}{\scriptsize{(-4.85)}}  & 73.68\textcolor{CustomGreen}{\scriptsize{(-1.20)}}\\
        Gradient Magnitude                               & 69.83\textcolor{CustomRed}{\scriptsize{(+3.00)}}  & 73.71\textcolor{CustomRed}{\scriptsize{(+2.82)}}  & 61.93\textcolor{CustomGreen}{\scriptsize{(-4.90)}}  & 69.88\textcolor{CustomGreen}{\scriptsize{(-1.01)}}\\
        Training Loss                                    & 70.11\textcolor{CustomRed}{\scriptsize{(+3.28)}}  & 74.54\textcolor{CustomRed}{\scriptsize{(+3.65)}}  & 59.46\textcolor{CustomGreen}{\scriptsize{(-7.37)}}  & 68.98\textcolor{CustomGreen}{\scriptsize{(-1.91)}}\\
        Feature Similarity \& Gradient Magnitude         & 71.63\textcolor{CustomRed}{\scriptsize{(+4.80)}}  & 75.56\textcolor{CustomRed}{\scriptsize{(+4.67)}}  & 64.31\textcolor{CustomGreen}{\scriptsize{(-2.52)}}  & 69.90\textcolor{CustomGreen}{\scriptsize{(-0.99)}}\\
        Feature Similarity \& Training Loss              & 71.86\textcolor{CustomRed}{\scriptsize{(+5.03)}}  & 75.98\textcolor{CustomRed}{\scriptsize{(+5.09)}}  & 63.28\textcolor{CustomGreen}{\scriptsize{(-3.55)}}  & 69.26\textcolor{CustomGreen}{\scriptsize{(-1.63)}}\\
        Prediction Similarity \& Gradient Magnitude*     & 73.32\textcolor{CustomRed}{\scriptsize{(+2.73)}}  & 78.23\textcolor{CustomRed}{\scriptsize{(+3.35)}}  & 65.34\textcolor{CustomGreen}{\scriptsize{(-5.25)}}  & 72.96\textcolor{CustomGreen}{\scriptsize{(-1.92)}}\\
        Prediction Similarity \& Training Loss*          & \textbf{73.65}\textcolor{CustomRed}{\scriptsize{(+3.06)}}  & \textbf{78.67}\textcolor{CustomRed}{\scriptsize{(+3.79)}}  & 64.88\textcolor{CustomGreen}{\scriptsize{(-5.71)}}  & 72.94\textcolor{CustomGreen}{\scriptsize{(-1.94)}}\\
        \bottomrule
    \end{tabular}
\end{table*}

\subsection{Comparison with Different Pacing Functions}
\begin{figure}[t]
	\centering
	\includegraphics[width=0.8\linewidth]{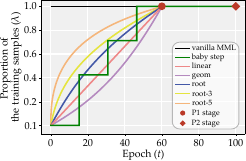}
        \captionsetup{font=normalsize} 
        \captionsetup{width=1\linewidth} 
        \caption{Visualization of different pace functions. The horizontal axis represents the training epoch number, while the vertical axis indicates the proportion of the most balanced samples available at the $t$-th epoch.}
        \label{fig:f2}
\end{figure}

In this subsection, we investigate the effect of different pacing functions as heuristic schedulers on imbalanced multi-modal learning. We consider various pacing functions, namely baby step, linear, root, root-3, root-5, and geometric. The visualization of these three pacing functions is presented in Figure~\ref{fig:f2}. The baby step function sorts data into bins from easy to hard, starting with the easiest and progressively adding harder bins after fixed epochs; the linear function increases training samples at a constant rate; root functions prioritize introducing harder samples within fewer epochs, while the geometric function dedicates more epochs to training on the easiest samples. For simplicity, we use prediction similarity as the measurer, setting the curriculum period $T_{grow}$ and the initial sample proportion $\lambda_0$ to 40 and 0.1, respectively. According to the performance on the CREMA-D dataset shown in Table~\ref{tab:tab3}, we observe that the root functions and the geometric function show a slight advantage, while vanilla training is inferior to all pacing functions. This demonstrates that an easy-to-hard training strategy helps the model to establish a robust feature representation capability, thus effectively mitigating modality imbalance.

\subsection{Sensitivity to Hyper-Parameters}
\begin{figure}[htbp]
	\centering
	\includegraphics[width=1\linewidth]{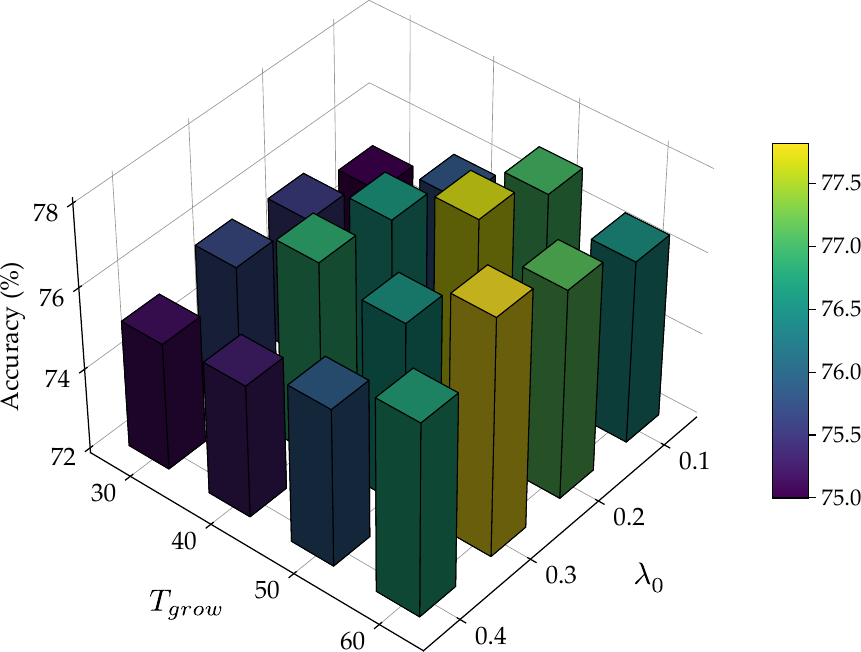}
        \captionsetup{font=normalsize} 
        \captionsetup{width=1\linewidth} 
        \caption{Parameter sensitivity analysis under the CL setting.}
        \label{fig:f3}
\end{figure}

We further examine the impact of the hyper-parameters $\lambda_0$ and $T_{grow}$ on MML performance. $\lambda_0$ determines the initial proportion of training samples, while $T_{grow}$ specifies the training epoch when the pacing function first reaches 1. For simplicity, prediction similarity is used as the measurer on the CREMA-D dataset, with $\lambda_0$ and $T_{grow}$ varied across \{0.1,0.2,0.3,0.4\} and \{30,40,50,60\}, respectively. 
We present the results in Figure~\ref{fig:f3}, which reveal the following: (1). Generally, as $\lambda_0$ increases, MML performance initially improves and then declines. Specifically, MML performance is optimal when $\lambda_0$ is 0.2 or 0.3. A very small $\lambda_0$ leads to insufficient training samples in the early stage. Conversely, an excessively large $\lambda_0$ introduces more hard samples, degrading MML performance. (2). Compared to $\lambda_0$, $T_{grow}$ is more sensitive, as reflected in its fluctuation. Hence, without extensive trials, identifying the optimal pacing function and its corresponding hyper-parameters for a specific task becomes challenging.

\begin{table}[t]
    \centering
    \captionsetup{font=normalsize} 
    \captionsetup{width=1\linewidth} 
    \caption{Comparison of different pacing functions under the CL setting.}
    \label{tab:tab3}
    \setlength{\tabcolsep}{3.5pt} 
    \begin{tabular}{lcc}
        \toprule
        \multirow{2}{*}{Pacing Functions} & \multicolumn{2}{c}{ACC\,(\%)}\\
        \cmidrule(lr){2-3}
                                                         & P1 (CL) & P2 (CL)\\
        \midrule
        Vanilla Training                                 & 70.59  & 74.88\\
        Baby Step                                        & 72.45\textcolor{CustomRed}{\scriptsize{(+1.86)}}  & 76.34\textcolor{CustomRed}{\scriptsize{(+1.46)}}\\
        Linear                                           & 72.18\textcolor{CustomRed}{\scriptsize{(+1.59)}}  & 76.88\textcolor{CustomRed}{\scriptsize{(+2.00)}}\\
        Root                                             & 73.54\textcolor{CustomRed}{\scriptsize{(+2.95)}}  & 77.12\textcolor{CustomRed}{\scriptsize{(+2.24)}}\\
        Root-3                                           & 73.66\textcolor{CustomRed}{\scriptsize{(+3.07)}}  & \textbf{77.55}\textcolor{CustomRed}{\scriptsize{(+2.67)}}\\
        Root-5                                           & 72.96\textcolor{CustomRed}{\scriptsize{(+2.37)}}  & 76.75\textcolor{CustomRed}{\scriptsize{(+1.87)}}\\
        Geometric                                        & \textbf{73.79}\textcolor{CustomRed}{\scriptsize{(+3.20)}}  & 77.42\textcolor{CustomRed}{\scriptsize{(+2.54)}}\\
        \bottomrule
    \end{tabular}
\end{table}

\section{Limitations and Future Work}
Our proposed method primarily addresses the modality imbalance problem in classification tasks. For other multi-modal downstream tasks, such as cross-modal retrieval and image captioning, can this imbalance phenomenon be mitigated from the perspective of training sequences? We leave this question for future work.

\bibliographystyle{named}
\bibliography{ijcai24}

%% file: defs.tex
\def\b{{\boldsymbol b}}

\def\e{{\boldsymbol e}}

\def\W{{\boldsymbol W}}

\def\X{{\boldsymbol X}}
\def\x{{\boldsymbol x}}
\def\Y{{\boldsymbol Y}}
\def\y{{\boldsymbol y}}

\def\z{{\boldsymbol z}}

\def\LM{{\mathcal L}}

\def\RB{{\mathbb R}}

\def\softmax{\mathtt{softmax}}

\def\Norm{\mathtt{Norm}}
\def\simm{\mathtt{sim}}
\def\min{\mathtt{min}}
\def\RandomSampling{\mathtt{Random\;Sampling}}
\def\SequentialSampling{\mathtt{Sequential\;Sampling}}